\documentclass[runningheads]{llncs}
\usepackage{graphicx}
\usepackage[breaklinks,colorlinks,citecolor=blue,linkcolor=blue,bookmarks=false]{hyperref}
\usepackage{xspace}
\usepackage{amsmath,amssymb}
\usepackage[normalem]{ulem}
\usepackage{xcolor,colortbl}
\usepackage{booktabs}
\usepackage{multirow}

\input{custom}
\renewcommand{\paragraph}[1]{\par\medskip\noindent\textbf{#1}\space}

\begin{document}

\title{Multimodal Learning for Embryo Viability Prediction in Clinical IVF}

\author{Junsik Kim\inst{1} 
\and
Zhiyi Shi\inst{2} 
\and
Davin Jeong\inst{1} 
\and
Johannes Knittel\inst{1} 
\\Helen Y. Yang\inst{1} 
\and
Yonghyun Song\inst{1} 
\and
Wanhua Li\inst{1} 
\and
Yicong Li\inst{1} 
\\Dalit Ben-Yosef\inst{4} 
\and
Daniel Needleman\inst{1,3} 
\and
Hanspeter Pfister\inst{1} 
}
\authorrunning{J. Kim et al.}
\institute{
    Harvard University, Cambridge MA 02138, USA
    \and
    Carnegie Mellon University, PA 15213, USA
    \and
    Flatiron Institute, New York, NY 10010, USA
    \and
    Tel Aviv Sourasky Medical Center, Tel Aviv, Israel
    \\
    \email{jskim@seas.harvard.edu}
}

\maketitle              

\begin{abstract}

In clinical In-Vitro Fertilization (IVF), identifying the most viable embryo for transfer is important to increasing the likelihood of a successful pregnancy. Traditionally, this process involves embryologists manually assessing embryos' static morphological features at specific intervals using light microscopy. This manual evaluation is not only time-intensive and costly, due to the need for expert analysis, but also inherently subjective, leading to variability in the selection process. To address these challenges, we develop a multimodal model that leverages both time-lapse video data and Electronic Health Records (EHRs) to predict embryo viability. 
One of the primary challenges of our research is to effectively combine time-lapse video and EHR data, owing to their inherent differences in modality. We comprehensively analyze our multimodal model with various modality inputs and integration approaches. Our approach will enable fast and automated embryo viability predictions in scale for clinical IVF.

\end{abstract}

\section{Introduction}

Infertility affects approximately one in six couples globally~\cite{cui2010mother}, propelling many towards assisted reproductive technologies such as In-Vitro Fertilization (IVF). 
IVF entails stimulating patients to produce multiple oocytes, which are then retrieved, fertilized in vitro, and the resultant embryos cultured. Selected embryos are transferred to the maternal uterus to initiate pregnancy, with surplus viable embryos cryopreserved for future attempts. Although transferring multiple embryos might increase the likelihood of conception, it simultaneously elevates the risk of multiple pregnancies, which are linked to heightened maternal and neonatal morbidity and mortality~\cite{norwitz2005maternal}. Consequently, there is a pressing need to limit embryo transfer to a single, optimally selected embryo to maximize the chances of a healthy singleton birth~\cite{lee2016elective} which remains challenging~\cite{racowsky2011national}. 

The prevailing practice in embryo selection primarily relies on morphological analysis through microscopic imaging. Embryos undergo a series of developments post-fertilization, transitioning through stages from pronuclei alignment to blastocyst formation, with clinicians traditionally scoring embryos based on discrete, manually observed morphokinetic features such as cell number, cell shape, cell symmetry, the presence of cell fragments, and blastocyst appearance~\cite{elder2000vitro}. 
Nowadays, many clinics adopt time-lapse microscopy incubators to capture movies of embryos continuously without disturbing their culture conditions~\cite{armstrong2019time}.
Despite this advancement, the analysis of these videos remains manual, which is labor-intensive and subjective.

Numerous studies have focused on predicting and analyzing the morphological features of embryos using images or videos, covering aspects like blastocyst size~\cite{kheradmand2017inner}, blastocyst grade~\cite{filho2012method,khosravi2019deep,kragh2019automatic}, cell boundaries~\cite{rad2018hybrid,jang2023amodal}, cell counting~\cite{khan2016deep,lau2019embryo}, and developmental stage prediction~\cite{lukyanenko2021developmental}.
Subsequently, a comprehensive pipeline employing deep learning models was developed to predict five key morphological features of embryos~\cite{leahy2020automated}, yielding outputs in the forms of classification, regression, and segmentation. 
These key morphological features are shown to be correlated to the live birth result of IVF treatments when converted to interpretable features by heuristic post-processing~\cite{yang2024blastassist}, such as the timing of stage transitions, cell symmetry index, and zona thickness. 
However, solely relying on the converted features may overlook more intricate and nuanced details of embryo development captured in videos. 
Additionally, these approaches mainly focus on visual features from time-lapse imaging and do not integrate data from Electronic Health Records (EHRs), which contain important variables such as patients' health information and treatment details.

In this work, we introduce a multimodal model for predicting embryo viability, leveraging both time-lapse videos and Electronic Health Records (EHRs). 
Although there has been an attempt to utilize image and EHR modalities~\cite{liu2023development}, their focus is not on multimodal integration, and they do not use video data.
A major challenge in multimodal learning is the effective integration of diverse modal types to ensure balanced training without modality bias~\cite{wang2020makes}. We explore different multimodal integration methods.
Inspired by the previous works, our multimodal model not only incorporates time-lapse videos and EHR data but also includes morphological~\cite{leahy2020automated} and interpretable features~\cite{yang2024blastassist} as additional inputs.
Through comprehensive experiments with diverse combinations of modalities, we analyze different multimodal integrations and demonstrate the effectiveness of our multimodal model for embryo viability prediction in clinical IVF.

\section{Dataset}

We collected data from 3,695 IVF treatment cycles with 24,027 embryos imaged every 20 minutes up to the first five days of development where each image size is $500\times500$. This corresponds to approximately 6 million images of embryos.
Additionally, electronic health record (EHR) data, including patient information, treatment information, and live birth records as a treatment outcome, are collected. 
Among the collected data samples, we curate the multimodal dataset with embryos that have both video and EHR modalities with treatment outcomes. 
Our multimodal dataset consists of a total of 1700 treatment cycles with 3318 embryos. Out of 1700 treatments, 260 treatments are successful with equal or more than one live birth.
It's important to note that each treatment cycle fertilizes multiple embryos, and only healthy embryos are selected for transfer. Some cycles freeze all embryos for future use rather than immediate transfer. Therefore, the number of embryos that have the treatment outcome is limited compared to the scale of the raw data collected.

\section{Method}

\begin{figure}[t]
    \centering
    \includegraphics[width=1.0\linewidth]{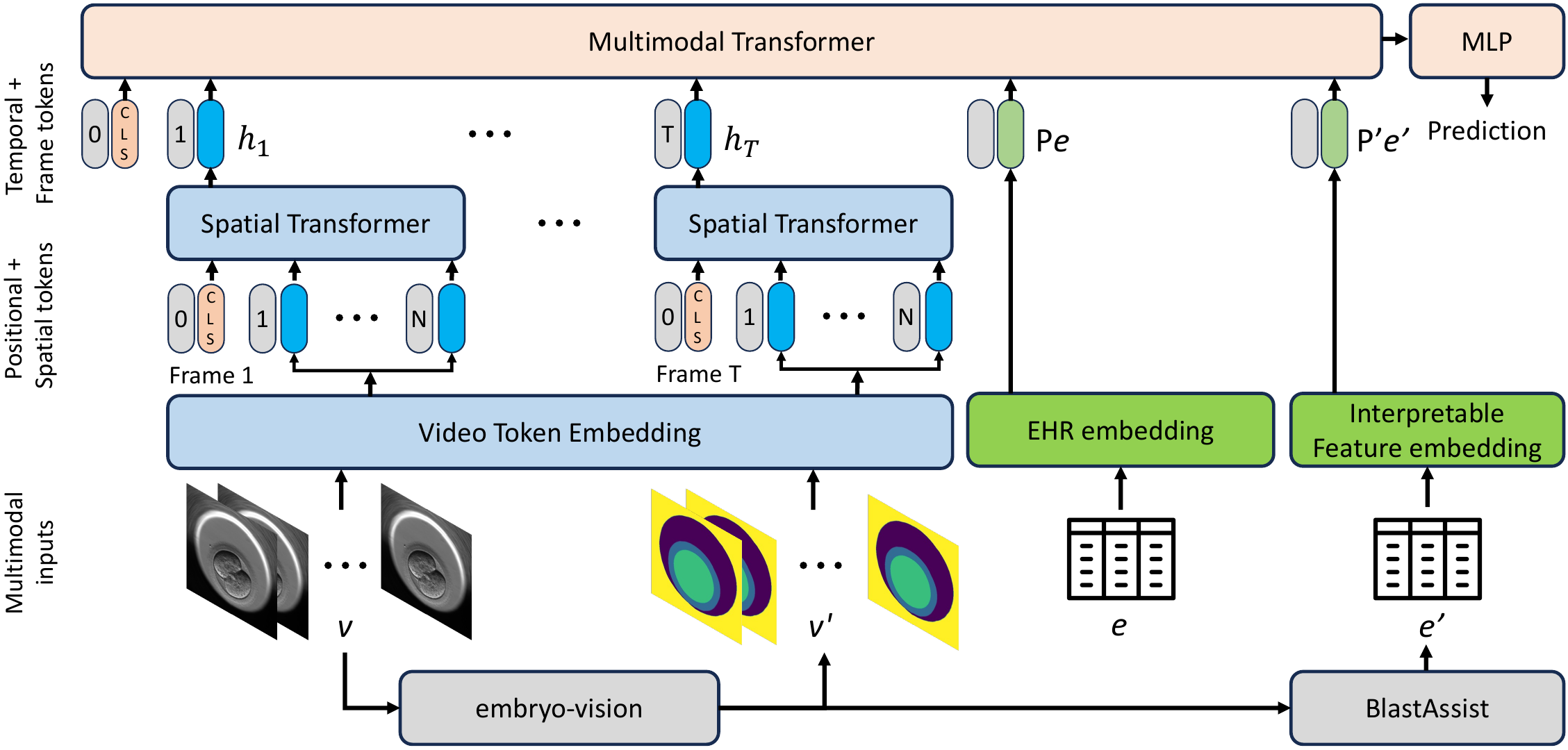}
    \caption{Overview of our multimodal model. Video data is first tokenized into patches per frame. Then, the spatial transformer encodes per frame embeddings. The Multimodal transformer inputs both frame embeddings and an EHR embedding to output a multimodal feature. Lastly, the MLP head predicts embryo viability based on the multimodal feature. If additional inputs in the form of video or tabular are available, such as outputs from Embryo-vision~\cite{leahy2020automated} or BlastAssist~\cite{yang2024blastassist}, they are processed in a similar manner as the video input and the EHR input respectively.}
    \label{fig:architecture}
\end{figure}

\begin{figure}[t]
    \centering
    \includegraphics[width=1.0\linewidth]{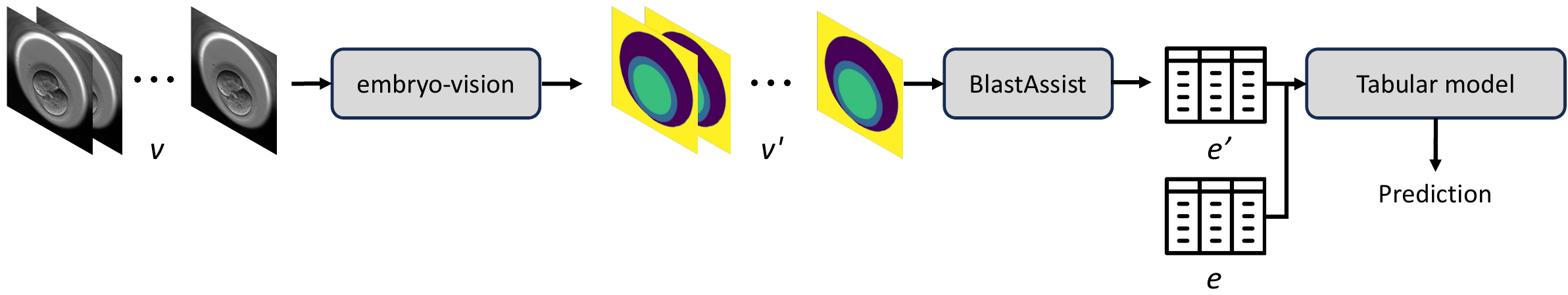}
    \caption{Overview of the two-stage approach. First, morphological features $\mathbf{v}'$ are extracted from videos using~\cite{leahy2020automated}. Then, the extracted features $\mathbf{v}'$ are converted to interpretable features $\mathbf{e'}$ in tabular format using~\cite{yang2024blastassist}. Lastly, the tabular model inputs EHRs $\mathbf{e}$ and interpretable features $\mathbf{e'}$ to predict embryo viability.
    }
    \label{fig:two_stage}
\end{figure}

In this work, we explore two different directions to integrate multimodal data for embryo viability prediction. One is a transformer-based multimodal model where EHRs and videos are processed end-to-end, as shown in~\Fref{fig:architecture}.
Another approach is to take a two-stage approach where the video data is first processed to extract morphological features in tabular format using off-the-shelf methods~\cite{leahy2020automated,yang2024blastassist}, and then input to the tabular models with EHRs as shown in~\Fref{fig:two_stage}. 
Although the two-stage approach can be modeled by a single tabular modality model, it is multimodal by nature as video data is converted and included in a tabular format.

Although there are several multimodal transformer models~\cite{sun2019videobert,akbari2021vatt,lin2021exploring,nagrani2021attention} available, it is not straightforward to apply them to embryo viability prediction as they assume samples in each modality has one-to-one correspondence. In our case, videos are embryo-specific, but EHRs are treatment-specific, which is shared across embryos within the same treatment cycle. Therefore it is difficult to directly apply cross-modal correspondence or contrastive learning as in other multimodal learning approaches. To this end, we propose a multimodal transformer that is based on a video transformer architecture with modifications to allow multimodal inputs.

\paragraph{Input modalities}
Let $\mathcal{T}_{n}= \{v_n, e_n\}$ be a multimodal sample in $n$-th treatment cycle in our multimodal dataset, where ${v}_{n}^{m} \in \mathbb{R}^{T \times H \times W \times C}$ denotes a time-lapse video of $m$-th embryo fertilized in $n$-th treatment cycle and ${e}_{n} \in \mathbb{R}^{C}$ denotes an EHR containing information of the patient and treatment applied. 
Note that time-lapse videos are embryo-specific, but EHR data corresponds to the treatment cycle; thus, they are not embryo-specific.
Our goal is to predict embryo viability formulated as 
$y = \frac{\mathrm{n\_births}}{\mathrm{n\_transferred}}$, where viability is defined as the number of births over the number of embryos transferred. 
The number of embryos transferred at each treatment cycle varies depending on various factors, such as the number of embryos fertilized, embryo quality examined by embryologists, or the patient's medical history. 

Other than video data, we can additionally utilize morphological embryo features extracted from videos by off-the-shelf methods, \eg, Embryo-vision~\cite{leahy2020automated}, and BlastAssist~\cite{yang2024blastassist}.
Embryo-vision outputs a set of features $v_{n,t}^{'m}$ from a video frame $v_{n,t}^{m}$, which are zona semantic segmentation $\mathbf{s}_z$, blastomere instance segmentation $\mathbf{s}_b$, pronuclei instance segmentation $\mathbf{s}_p$, fragmentation regression $\mathbf{r}$, and stage classification $\mathbf{c}$. BlastAssist further converts the morphological features into a set of interpretable features $\mathbf{e'}$ such as zona well thickness, stage transition timing, and cell symmetry index. For more details, refer to the supplementary and~\cite{leahy2020automated,yang2024blastassist}.

\paragraph{Video transformer}
Videos are significantly larger than the size of other modalities. Directly applying spatio-temporal attention to a video would result in a large number of tokens, which require an immense amount of memory and computation. Inspired by ViViT~\cite{arnab2021vivit}, we design a transformer in a factorized encoder structure where spatial attention is applied first, followed by temporal attention. 

For spatial attention, we first tokenize each frame $v^{m}_{n,t} \in \mathbb{R}^{H \times W \times C}$ to a set of tokens by extracting non-overlapping image patches $x_i \in \mathbb{R}^{h \times w \times C}$ and then apply linear projection $\mathbf{E}$. 
Then, a set of embedded frame tokens and a learnable class token are added to a learnable positional embedding $\mathbf{p}$ and passed through a transformer consisting of a sequence of $L$ transformer layers to output a frame-level representation. 
\begin{equation}
    \mathbf{z} = [z_{\text{cls}}, \mathbf{E}x_1, \ldots, \mathbf{E}x_N] + \mathbf{p}
    \label{eq:spatial_emb}
\end{equation}
Each transformer layer $\ell$ consists of Multi-Headed Self-Attention~\cite{vaswani2017attention}, layer normalisation(LN)~\cite{ba2016layer}, and MLP blocks as follows:
\begin{align}
   \mathbf{y}^{\ell}=\mathrm{MSA}(\mathrm{LN}(\mathbf{z}^\ell))+\mathbf{z}^\ell\\
    \mathbf{z}^{\ell+1}=\mathrm{MLP}(\mathrm{LN}(\mathbf{y}^\ell))+\mathbf{y}^\ell
\end{align}
The output token $z_{cls}^{L}$ embeds frame-level representation. 
Temporal attention is performed similarly to spatial embedding by applying $L'$ transformer layers on a set of frame tokens $\mathbf{h}$,
\begin{equation}
    \mathbf{h} = [h_{\text{cls}}, z_{cls,1}^{L}, \ldots, z_{cls,T}^{L}] + \mathbf{t}
    \label{eq:temporal_emb}
\end{equation}
where $h_{cls}$ is a learnable class token in temporal attention, and $\mathbf{t}$ is a learnable temporal embedding.

\paragraph{Multimodal Transformer}
We modify a video transformer to allow multimodal inputs.
We embed EHR data $\mathbf{e}$ by linear projection and then append to the frame tokens. If we have additional features in a tabular format, \eg, interpretable features $\mathbf{e'}$, it is processed in the same way as EHR data.
With EHR data tokens, the temporal attention input in~\Eref{eq:temporal_emb} becomes multimodal attention input as follows,
\begin{equation}
    \mathbf{h} = [h_{\text{cls}}, h_1, \ldots, h_T, \mathbf{P}e, \mathbf{P'}e'] + \mathbf{t}
\end{equation}
where $h_t$ is a frame token at frame $t$,  $\mathbf{P}$ and $\mathbf{P'}$ are linear projections for $\mathbf{e}$ and  $\mathbf{e'}$ respectively.
When only video is input to the model, a frame token $h_t$ becomes $z_{cls,t}^L$ as in~\Eref{eq:temporal_emb}. 
Additionally, we can incorporate more per-frame modality inputs from Embryo-vision to enrich the representation of a frame token $h_t$. 
The Embryo-vision outputs a set of morphological features 
$v' =\{ s_z, s_b, s_p, r, c\}$ 
where the first three features are segmentation masks and the latter two are vectors. 
The mask format features are passed to the spatial attention and processed similarly to the video input. For simplicity, let's denote spatial transformer operation as $f_s: \mathbb{R}^{H \times W \times C} \rightarrow \mathbb{R}^d $. When a video is input, $f_s(v_t)$ equals $z_{cls,t}^L$ as in~\Eref{eq:temporal_emb}. When multiple video modalities are available, the frame token $h_t$ is formulated as a concatenation of tokens from different modalities as follows,
\begin{align}
    h_t = [f_s(v_t), f_s(s_{z,t}), f_s(s_{b,t}), f_s(s_{p,t}), \mathbf{E'}[r_t,c_t]]
\end{align}
where $\mathbf{E'}$ is a linear projection applied to the concatenation of $r_t$ and $c_t$.

\section{Experiments}

\paragraph{Implementation details}
For spatial attention, we use the pre-trained DeiT-Ti~\cite{touvron2021training} as a spatial transformer without fine-tuning. We attempted to fine-tune a spatial transformer, but it resulted in worse performance due to the limited number of labeled samples. 
For temporal or multimodal attention, we use 4 transformer layers. Input frames are resized from $500 \times 500$ to $224 \times 224$. Videos are clipped to have a maximum of 360 frames since this corresponds to the first 5 days of observation, where each frame is captured at 20-minute intervals. To enable memory-efficient training, we subsample every 4 frames, resulting in 90 frames per video.
Flip and rotation are applied to videos and masks for augmentation. The batch size is set to 4, the learning rate is set to 1e-4, and the model is trained until the validation loss converges.
MLP head consists of two fully connected layers with ReLU activation in between. Huber loss~\cite{huber1992robust} is used to train the multimodal transformer. The experiments are performed using one A100 GPU.

\begin{table}[t]
\centering
\caption{Number of successful and failed treatments and embryos in each split in the form of ``number of embryos" / ``number of treatments."
}
\label{tab:ehr}

\begin{tabular}{@{}llll@{}}
\toprule
Split & Total & Success & Fail \\ 
\midrule
Train   & 2617/1360  & 362/208   & 2255/1152  \\
Val     & 327/170    & 54/26     & 273/144    \\
Test    & 342/170    & 54/26     & 288/144    \\
\bottomrule
\end{tabular}

\end{table}

\paragraph{Experiment setup}
We randomly split train, validation, and test splits to an 8:1:1 ratio while preserving the success rate within each split. For evaluation, we use two performance metrics: the area under the receiver operating characteristic curve (ROCAUC) and F1-Score. We evaluate two different scenarios: embryo viability prediction and treatment success prediction. Each treatment has equal to or more than one embryo transferred. In the embryo viability prediction scenario, we set the ground truth label to '1' for all embryos transferred (instead of $\mathrm{\frac{n\_births}{n\_transferred}}$) if the treatment is successful, then compute AUCROC and F1-Score. In treatment success prediction, we sum the viability predictions of embryos transferred together and then calculate AUCROC and F1-Score. For F1-Score measurement, we use 0.15\footnote{The treatment cycle with the highest number of embryos transferred is 5. Therefore, embryo viability values in successful treatments range from 0.2 to 1.0.} as a threshold for embryo viability prediction and 0.5\footnote{Treatment success is defined as the n\_births value equal to or higher than 1.} for treatment success prediction. 
F1-Score quantifies the precision of predictions at a fixed threshold, whereas AUCROC measures capability in assessing the relative quality of the samples.

\paragraph{Two-stage approach}
We compare our multimodal transformer with two-stage approaches using two transformer based methods: TabTransformer~\cite{huang2020tabtransformer} and TabNet~\cite{arik2021tabnet}. We follow the official implementation of~\cite{borisov2022deep}~\footnote{\url{https://github.com/kathrinse/TabSurvey}} to train tabular models with the best hyperparameters after performing hyperparameter search using cross-validation. For more details, refer to the supplementary.

\begin{table}[t]
    \centering
    \caption{Performance comparison on embryo viability prediction with different modalities using a multimodal transformer. \textbf{v} is a video modality, \textbf{v'} is an output from Embryo-vision, \textbf{e} is EHR data, and \textbf{e'} is an output from BlastAssist. The best performance is marked in bold.
    }
    {
		\begin{tabular}{llcc@{\hspace{20pt}}cc}
            \toprule
            Modality &  \multicolumn{2}{c}{Embryo} & \multicolumn{2}{c}{Treatment}\\
             & AUCROC & F-1 & AUCROC & F-1 \\
            \midrule
            \multirow{1}{*}{\textbf{v}}          & 0.578 & 0.284 & 0.579 & 0.315 \\
            \multirow{1}{*}{\textbf{v+e}}        & 0.580 & 0.297 & 0.581 & 0.286\\ 
            \multirow{1}{*}{\textbf{v+v'}}       & 0.676 & 0.316 & 0.675 & \textbf{0.336}\\
            \multirow{1}{*}{\textbf{v+v'+e+e'}}  & 0.647 & 0.296 & 0.643 & 0.310 \\
            \midrule
            \multirow{1}{*}{\textbf{v'}}         & 0.666 & 0.317 & \textbf{0.697} & 0.313\\
            \multirow{1}{*}{\textbf{v'+e+e'}}    & \textbf{0.688} & \textbf{0.338} & 0.683 & 0.312\\
            \bottomrule
		\end{tabular}
    }
    \label{tab:multimodal}
\end{table}
\begin{table}[ht]
    \centering
    \caption{Performance comparison on embryo viability prediction with different modalities using a two-stage approach. \textbf{e} is EHR data, and \textbf{e'} is an output from BlastAssist. Confidence intervals are reported with 10 runs.}
    {
    \resizebox{1.0\linewidth}{!}{
		\begin{tabular}{llcc@{\hspace{20pt}}cc}
            \toprule
            Modality & Method & \multicolumn{2}{c}{Embryo} & \multicolumn{2}{c}{Treatment}\\
            & & AUCROC & F-1 & AUCROC & F-1 \\
            \midrule
            \multirow{2}{*}{\textbf{e}} 
                & TabTransformer~\cite{huang2020tabtransformer} & 0.586 $\pm$ 0.045 & 0.110 $\pm$ 0.068 
                                                                & 0.604 $\pm$ 0.054 & 0.167 $\pm$ 0.111 \\
                & TabNet~\cite{arik2021tabnet}                  & 0.591 $\pm$ 0.016 & 0.240 $\pm$ 0.020
                                                                & 0.631 $\pm$ 0.017 & 0.113 $\pm$ 0.033 \\
            \midrule
            \multirow{2}{*}{\textbf{e+e'}} 
                & TabTransformer~\cite{huang2020tabtransformer} & 0.634 $\pm$ 0.025 & 0.298 $\pm$ 0.045
                                                                & 0.681 $\pm$ 0.023 & 0.100 $\pm$ 0.031\\
                & TabNet~\cite{arik2021tabnet}                  & 0.629 $\pm$ 0.025 & 0.244 $\pm$ 0.042 
                                                                & 0.672 $\pm$ 0.026 & 0.188 $\pm$ 0.058\\
            \midrule
            \multirow{2}{*}{\textbf{e'}} 
                & TabTransformer~\cite{huang2020tabtransformer} & 0.593 $\pm$ 0.021 & 0.235 $\pm$ 0.040
                                                                & 0.624 $\pm$ 0.022 & 0.134 $\pm$ 0.030 \\
                & TabNet~\cite{arik2021tabnet}                  & 0.623 $\pm$ 0.012 & 0.232 $\pm$ 0.042
                                                                & 0.630 $\pm$ 0.023  & 0.146 $\pm$ 0.045 \\
            \bottomrule
		\end{tabular}
    }
    }
    \vspace{-0.5em}
    \label{tab:two_stage}
\end{table}

\paragraph{Experiments with multimodal transformer}
We evaluate our multimodal transformer on embryo viability prediction task using different combinations of modalities in~\Tref{tab:multimodal}.
The first 4 rows in the table show the results with video modality. 
The model trained with only video modality performs worse than the other modality combinations. When both video and EHR modalities are used, AUCROC marginally improves.
On the other hand, the model performance improves significantly when semantic features are added.
This shows that directly predicting embryo viability is challenging and semantic information is important for the prediction.
However, adding tabular format modalities to video modalities did not improve the prediction.
We conjecture this is due to the increased complexity of multimodal data to learn given limited training samples.
The performance drop with interpretable features is noticeable with video modality, but the performance drop is not observed in other combinations of modalities.

We evaluate the multimodal model without a video input $v$ in the last 2 rows in~\Tref{tab:multimodal}. The results without a video modality perform better than those with a video modality. 
This may be due to the limited number of training videos to learn good representation. We deploy a pre-trained vision transformer DeiT-Ti~\cite{touvron2021training} to overcome the limited training set size, but multimodal transformer layers are trained from scratch; therefore, the multimodal attention is performed in a sub-optimal way. 
On the contrary, a model trained with Embryo-vision outputs $v'$ performs significantly better than those with $v$. Unlike raw video, Embryo-vision outputs are in the form of segmentation masks, which are semantically meaningful and have a simple visual structure. Therefore, it is easier for the model to understand and optimize the weights to extract relevant features for the task.

\paragraph{Experiments with two-stage approach}
We compare the two-stage approach with different types of tabular models. 
Unlike the end-to-end multimodal learning method, we observed higher performance variation in two-stage methods. We conjecture this is due to the early convergence of two-stage models, which results in different solutions.
Here, we report confidence intervals from 10 trials of the two-stage approaches. 
Among different modalities, using both EHR and interpretable features performs best for the two-stage approaches. 
Although visual data is not directly input to the model, interpretable features encode visual information; therefore, the tabular models show competitive performance when using both EHRs and interpretable features.

One noticeable difference to the multimodal transformer is the low F-1 score on treatment success prediction.
Although tabular models are trained with regression objectives, they fail to calibrate the prediction confidence, resulting in a low F-1 score.
In practice, finding the best threshold is a difficult problem. Therefore, without an appropriate threshold estimation method, a model with good confidence calibration is favored.
If an optimal threshold can be found, a higher F-1 score will be achieved for both multimodal transformers and two-stage tabular models.
\section{Discussion and Conclusion}
One challenge in multimodal learning with supervision is the size of the training data. Although the collected data is not on a small scale, the embryos with a treatment outcome are very limited. This hinders the supervised training of large-scale models. 
We conjecture the negative effect of video modality in~\Tref{tab:multimodal} is also due to the limited training size.
One solution is to pre-train modality-specific encoders separately with pretext tasks using self-supervised learning~\cite{ericsson2022self} and then fine-tune the encoders with multimodal transformers by supervised learning for the downstream task. 
With better encoder representations by self-supervised learning, a multimodal transformer will effectively integrate modality features without performance degradation.

In this work, we explore two approaches to incorporate time-lapse videos and EHRs to build a multimodal model for embryo viability prediction. First, we build a multimodal transformer to allow different modalities to be integrated together. The multimodal inputs include not only videos and EHRs but also other morphological features extracted from off-the-shelf methods. 
We also explore an alternative method; a two-stage approach where the first stage is to extract and convert visual morphological features to tabular format and then combine it with EHRs for tabular models. 
The experiments with various modalities demonstrate the effectiveness of our multimodal model over two-stage approaches. We further analyze which modality is important in predicting embryo viability.
In future research, we will explore pre-training and fine-tuning methodologies to address the challenge posed by the limited size of supervised training sets in multimodal learning.

\begin{credits}
\subsubsection{\ackname}
We thank all affiliates of the Harvard Visual Computing Group for their valuable feedback. This work was supported by NIH grant R01HD104969 and Harvard Data Science Initiative Postdoctoral Fellowship.

\subsubsection{\discintname}
The authors have no competing interests to declare that are relevant to the content of this article.
\end{credits}

%
%
\bibliographystyle{splncs04}
\bibliography{main.bbl}
\end{document}